\pdfoutput=1

\documentclass[11pt]{article}

\usepackage[preprint]{acl}

\usepackage{times}
\usepackage{latexsym}

\usepackage[T1]{fontenc}

\usepackage[utf8]{inputenc}

\usepackage{microtype}

\usepackage{inconsolata}

\usepackage{graphicx}

\usepackage{tasks}
\usepackage{svg}
\usepackage{booktabs} 
\usepackage{multirow}
\usepackage{amssymb} 
\usepackage{float} 

\title{A Structured Literature Review on Traditional Approaches in Current Natural Language Processing}

\author{Robin Jegan \and Andreas Henrich \\
        \\Otto-Friedrich-Universität Bamberg, Lehrstuhl für Medieninformatik\\
        \{robin.jegan,andreas.henrich\}@uni-bamberg.de}

\begin{document}
\maketitle
\begin{abstract}
The continued rise of neural networks and large language models in the more recent past has altered the natural language processing landscape, enabling new approaches towards typical language tasks and achieving mainstream success. Despite the huge success of large language models, many disadvantages still remain and through this work we assess the state of the art in five application scenarios with a particular focus on the future perspectives and sensible application scenarios of traditional and older approaches and techniques.

We survey recent publications in the application scenarios~\textit{classification},~\textit{information} and~\textit{relation extraction},~\textit{text simplification} as well as~\textit{text summarization}. After defining our terminology, i.e., which features are characteristic for ``traditional'' techniques in our interpretation for the five scenarios, we survey if such traditional approaches are still being used, and if so, in what way they are used. It turns out that all five application scenarios still exhibit traditional models in one way or another, as part of a processing pipeline, as a comparison/baseline to the core model of the respective paper, or as the main model(s) of the paper.
\end{abstract}
\section{Introduction}

The research field of natural language processing (NLP) has been transformed in recent years. Advancements in computing capabilities and research breakthroughs such as language models and since the end of 2022 large language models (LLMs) have impacted nearly every facet of NLP, reaching mainstream use through prompt-based solutions such as ChatGPT and other LLMs accessible via chat~\citep{stojkovic2024towards}. However, drawbacks remain despite the demonstrably impressive capabilities of LLMs, such as missing reproducibility, hallucinations, high resource requirements and others~\citep{yao2023llm,atil2024llm,stojkovic2024towards}. This structured literature review was started for two reasons: First, to gain an overview of techniques that LLMs replaced. Second, to gauge whether those replaced techniques, and even much older techniques, are still relevant in today's NLP landscape. 

Through this survey, a more thorough look at a few selected applications in text processing will analyze if these older, simpler and/or less resource-intensive techniques, models or approaches -- which we will call ``traditional'' from now on -- are still apparent in today's NLP papers. If such techniques exist, we will further examine their role in the respective application scenario. In short, traditional techniques comprise different approaches depending on the application scenarios, which we will detail below, e.g., extractive text summarization algorithms (in contrast to generative models) or rule-based approaches for information extraction. 

In this structured literature review, we look at five different application scenarios within NLP: 
\textit{classification}, \textit{information extraction}, \textit{relation extraction}, \textit{text simplification} and \textit{text summarization}.
The application scenarios will be analyzed according to three main research questions:
\begin{description}
\label{rq}
    \item[RQ1:] Are there still traditional, simple, and less resource-intensive techniques used in current research papers, for instance either as benchmark, as important part of a processing pipeline or as core method in specific application scenarios?
    \item[RQ2:] Why are the traditional models from RQ1 still being used and in what research field?
    \item[RQ3:] What are advantages of traditional models, if any, when compared to modern techniques?
\end{description}
The SLR will show that for each of the five application scenarios traditional models still exist in current research, in different use-cases and in varying quantities. All three purposes mentioned in~\textbf{RQ1} are still present in various papers, thus showing the still ongoing relevance of traditional models in today's NLP landscape~\footnote{See the complete statistics on \url{https://doi.org/10.5281/zenodo.13683800}.}.

Our paper is structured as follows: We will detail the related work in section~\ref{sec:rel} and afterwards the characteristics of traditional models in section~\ref{sec:term}. The methodology and quantitative results of the SLR will be the main focus of section~\ref{sec:slr}. Qualitative results of the survey and current research on approaches as mentioned above will be presented in section~\ref{sec:curr}, before typical application scenarios for traditional approaches are described in section~\ref{sec:appsec}. Section~\ref{sec:concl} concludes our paper.

\section{Related Work}
\label{sec:rel}
This section comprises an overview on current state-of-the-art approaches for the five application scenarios mentioned above. Our goal here is to present the state of the art for the five application scenarios in order to examine the prevalent models and approaches currently in use. 

\subsection{Classification}
Classification scenarios in NLP include clustering or categorization tasks, but generally refer to the process of a model labeling a text with a class value that is representative of a certain class of texts~\citep{aggarwal2012survey}. Traditionally, techniques such as support vector machines (SVM), decision trees and pattern- or rule-based classifiers were applied to solve this NLP task~\citep{aggarwal2012survey}. The mentioned traditional models were a small part of two surveys on text classification~\citep{li2022survey,kadhim2019survey}, the latter focusing on supervised techniques, followed by a survey on semi-supervised learning~\citep{duarte2023review}. 

The state of the art for text classification is presently dominated by either approaches using neural networks or language models~\citep{li2022survey}. Various neural network variants have been adapted for NLP applications, including architectures based on Convolutional Neural Networks (CNN), Graph Neural Networks (GNN) or Recurrent Neural Networks~\citep{minaee2021deep}.  Language models, at least prior to the end of 2022, were often using techniques such as BERT, XLNet, GPT or similar approaches~\citep{gasparetto2022survey}, all based on the transformer architecture~\citep{vaswani2017attention}. The state of the art has changed, however, with the advent of LLMs, which have upended leaderboards for various NLP application scenarios~\footnote{For example:~\url{https://huggingface.co/spaces/mteb/leaderboard}~(last accessed: 27.01.2025)}.

\subsection{Information Extraction}
Information extraction scenarios in NLP usually involve the identification and classification of entities in categories such as persons, temporal aspects, organizations and others~\citep{grishman2015information}. Rule-based approaches were traditionally used for this scenario, adapted for each entity type, or other approaches such as hidden Markov models or again SVMs~\citep{grishman2012information}.

Other tasks such as entity extraction for various entity types~\citep{ding2021few}, event extraction~\citep{hsu2021degree} and aspect-based sentiment analysis~\citep{lv2023efficient} can also be identified as information extraction. Again, most state-of-the-art approaches leverage the transformer architecture~\citep{ding2021few,lv2023efficient}. LLMs also have been studied within information extraction~\citep{li2023evaluating}, but still not completely surpassing the prior state-of-the-art for use-cases like aspect-oriented opinion extraction~\citep{han2023information}.

\subsection{Relation Extraction}
While closely connected to the previous NLP application, relation extraction requires a stronger focus on the connection between entities, usually characterized as semantic relations between them, whereas information extraction deals with more independent entities~\citep{zhang2017review,wang2023models}. Early approaches used similar techniques as information extraction applications, i.e., SVM once more, or co-training as a semi-supervised method~\citep{zhang2017review}. In general, relation extraction can be approached in various ways, as a classification or clustering task, sequence-to-sequence generation problem or sequence labeling task~\citep{detroja2023survey}. 

In recent years, a large number of approaches were applied to relation extraction, comprising architectures based on CNN, RNN, and GNN, or once again transformer-based approaches like BERT, and combinations of these techniques, in order to extract the relations between entities~\citep{detroja2023survey,sui2023joint,wang2023models}. Furthermore, there is an interplay between relation extraction techniques and structures such as knowledge graphs or graph networks~\citep{li2023reviewing,zhao2021representation}. LLMs have been studied for relation extraction as well, e.g., in the temporal domain~\citep{yuan2023zero}. Furtermore, LLMs achieve similar performance compared to the best fully supervised methods for relation extraction~\citep{wadhwa2023revisiting} .

\subsection{Text Simplification}
Text simplification scenarios in NLP transform a complex text into a simpler variant, while containing most of the content of the original text~\citep{chandrasekar1996motivations}. Text simplification, and in most cases text summarization as well, differs from the previously mentioned approaches in so far that a transformed text is produced, either by extraction or generation. For simplification, early approaches were applied in various ways, ranging from lexical substitutions~\citep{de2010text}, i.e., the replacement with simpler words using lexicons such as WordNet~\citep{miller1995wordnet}, to syntactic modifications~\citep{chandrasekar1996motivations}, by using finite state grammars in order to reduce the complexity of sentence structures. 

More modern approaches apply complex machine learning methods, which realize the simplification as a semantic or mono-lingual translation task, or even hybrids of lexical, syntactic and machine learning processes~\citep{al2021automated,janfada2020review}. Still, even in this more expansive NLP use-case, in terms of the amount of text that is produced compared to the previous scenarios, neural networks~\citep{alkaldi2023text,truicua2023simplex} and language models have taken over the state of the art~\citep{vasquez2023document,yamaguchi2023gauging}. The large majority of recent text simplification systems thus favor transformer-based approaches such as BART~\citep{anschutz2023language,sun2023teaching}, BERT~\citep{qiang2023unsupervised} or RoBERTa~\citep{cripwell2023document}.

\subsection{Text Summarization}
Text summarization scenarios in NLP also deal with producing a new version of some original text, however in a shortened form~\citep{hovy1998automated}. Extractive approaches, which identify and extract the unedited text from the source document, have to be contrasted with abstractive approaches, which generate some new output based on the source text~\citep{hovy1998automated}. Traditional methods for summarization have been largely based on statistical approaches, e.g., by using measures such as TF-IDF (term frequency-inverse document frequency), to identify the words, phrases or sentences, that should be selected for an extractive summary. Other traditional techniques include graph-based, topic-based or discourse-based approaches as well as early machine learning based techniques~\citep{gambhir2017recent}.

Two different scenarios have to be further separated: single-document and multi-document summarization~\citep{el2021automatic}. Multi-document summarization was approached using different neural networks such as CNNs, RNNs and GNNs, but the prevalent techniques in recent years were, once more, transformer-based and hybrid approaches~\citep{ma2022multi}. A stronger focus in this work, however, is placed on single-document summarization. The focus of most approaches that deal with single documents has been for some time on abstractive~\citep{li2024multilingual} or at least hybrid techniques~\citep{roy2024enhancing} and more recently also on LLMs~\citep{haldar2024analyzing,li2024improving}.

\section{Terminology}
\label{sec:term}
In order to conduct the SLR for our specific point of view, i.e., traditional approaches for the selected application scenarios, we have to specify the terminology of what we would categorize as ``traditional methods''. A clear cut and small definition is difficult since we look at diverse applications, but they nevertheless share some common features. 

One key aspect includes~\textit{reproducibility}. We look for approaches, that can produce the exact same output for a given input, prompt or query. LLMs and generative approaches in general are approaches that often suffer in this area and therefore, depending on the use-case, can lead to problems for evaluation or comparative studies. When running an LLM or other neural network yourself there are parameters in order to restrict more creative or diverse output, often called temperature, but restricting this parameter in order to produce repeatable outputs can lead to worse performance overall for the generated texts~\citep{peeperkorn2024temperature}. 

\textit{Efficiency} is another factor for traditional models. There are different facets, the first one being time. Execution time, meaning the time the model is running in order to produce the output, differs widely depending on the model. Traditional models, in general, were developed using far fewer computational resources, meaning the execution with today's capabilities usually completes much quicker as well. LLMs or neural networks, which often require dedicated GPUs in order to accelerate their training as well as execution, are much more time-intensive when compared to more traditional models~\citep{zhao2023survey}. 

Directly connected to the execution of the model is another dimension of efficiency, i.e., resources. Monetary costs in case of API requests as well as infrastructure or server costs can scale quickly with modern models, especially compared to more traditional models. Additionally, another more tangible part of resources, the required hardware, can differ here extremely as well. As mentioned above, dedicated GPUs are commonly used in order to accelerate the use of neural networks or LLMs, while traditional models can, in most cases, be executed on the CPU alone. Closely tied to this aspect is the energy-consumption of the system in use. Just as one sample, the International Energy Agency calculated at the beginning of 2024, that one API call to ChatGPT requires 2.9 Wh per request, a huge increase when compared to one Google search call (0.3 Wh per request)~\footnote{\url{https://iea.blob.core.windows.net/assets/6b2fd954-2017-408e-bf08-952fdd62118a/Electricity2024-Analysisandforecastto2026.pdf}~(last accessed: 27.01.2025)}.

One final factor is~\textit{documentation} as well as~\textit{explainability}. Traditional models, since they have existed for decades in many cases, have been widely used as well as implemented in many packages and programming languages. Documentation thus exists in various ways, ranging from technical documentation, tutorials to handbooks. Directly associated with documentation is their reduced complexity, especially when compared to neural networks and LLMs. Approaches such as extractive summarization approaches, for which groundbreaking publications date back to the 1950s~\citep{luhn1958automatic} and 1960s~\citep{edmundson1969new}, or decision trees, dating back to the 1980s~\citep{breiman2017classification}, define the processing steps in cases. These cases can be directly implemented as, for example, if-else-statements or simple sentence rankings, using e.g., TF-IDF, enabled through the invention of the IDF-measure in the 1970s~\citep{sparck1972statistical}.

When looking at our five application scenarios, rule-based techniques can be applied first and foremost for both extraction scenarios, i.e., information and relation extraction, but have traditionally been applied to many scenarios, e.g., anaphora resolution~\citep{hobbs1978resolving} or part-of-speech tagging~\citep{brill1992simple}. Rule-based techniques serve as the basis for decision trees, which can be selected as one traditional model for text classification as well as simplification approaches. SVMs, implemented for classification purposes in the 1990s~\citep{boser1992training}, and Naive-Bayes (NB), applied for the first time in text classification in the 1960s~\citep{mosteller1963inference}, are further traditional classification techniques. Random forest (RF) classifiers, originated in the 1990s~\citep{ho1995random}, can also be used to measure lexical complexity, which in turn can be used for simplification purposes. Finally, extractive approaches, sometimes analogous to rule-based techniques, would be considered as one traditional approach for text summarization.

\section{Structured Literature Review}
\label{sec:slr}

We have conducted our SLR based on a few assumptions. In order to ensure a certain scientific quality and at the same time limit the effort, we have used only one research database, the Association for Computing Machinery Digital Library (ACM DL). Thus, we can guarantee that nearly every retrieved publication has undergone a peer review process,~\footnote{About 80\% of publications in the ACM DL are full-length papers, that have been peer-reviewed, as per \url{https://libraries.acm.org/subscriptions-access/acmopen} (last accessed: 27.01.2025)} which would not have been the case if we had used other research databases such as Google Scholar or arXiv, where pre-print versions can be uploaded. In addition, the broad coverage of the ACM DL still ensures a certain degree of representativeness. We have further chosen the ACM DL due to its expansive library (over 3.8 Million publications (as of January 2025), in contrast to the much smaller ACL Anthology. The search functionality of the ACL Anthology is furthermore much more restrictive, in terms of advanced search capabilities. In the end our goal was not an all-encompassing view of the complete state-of-the-art, but rather a restricted view on high-quality publications, that could qualitatively be analyzed according to our research question.

Another pre-determined selection criterion was the time of publication, in our case the year 2023. In order to survey the impact of LLMs we decided to exclude newer publications, which could not be surveyed comprehensively due to the still-ongoing review and publication process for journals or conferences for 2024, that will be indexed to the ACM DL in the coming months. The results retrieved from the ACM DL were conducted using only the title of the paper, and further limited to matches for the exact search query, meaning, e.g., 'text classification' in that exact order.
In order to include text simplification in the SLR, the search criteria had to be expanded by querying the abstract as well due to the fact that no papers were retrieved by using the title alone. 

Furthermore, we recognize that the selection criteria for papers constrain us to a small, but in our opinion, representative snapshot of the current research in NLP in the chosen application scenarios. Thus, we hope to gain knowledge about the state of current NLP research, especially in terms of how traditional models are still relevant and are still in use.

\begin{figure}[hbtp]
\centering
\includegraphics[width=0.48\textwidth]{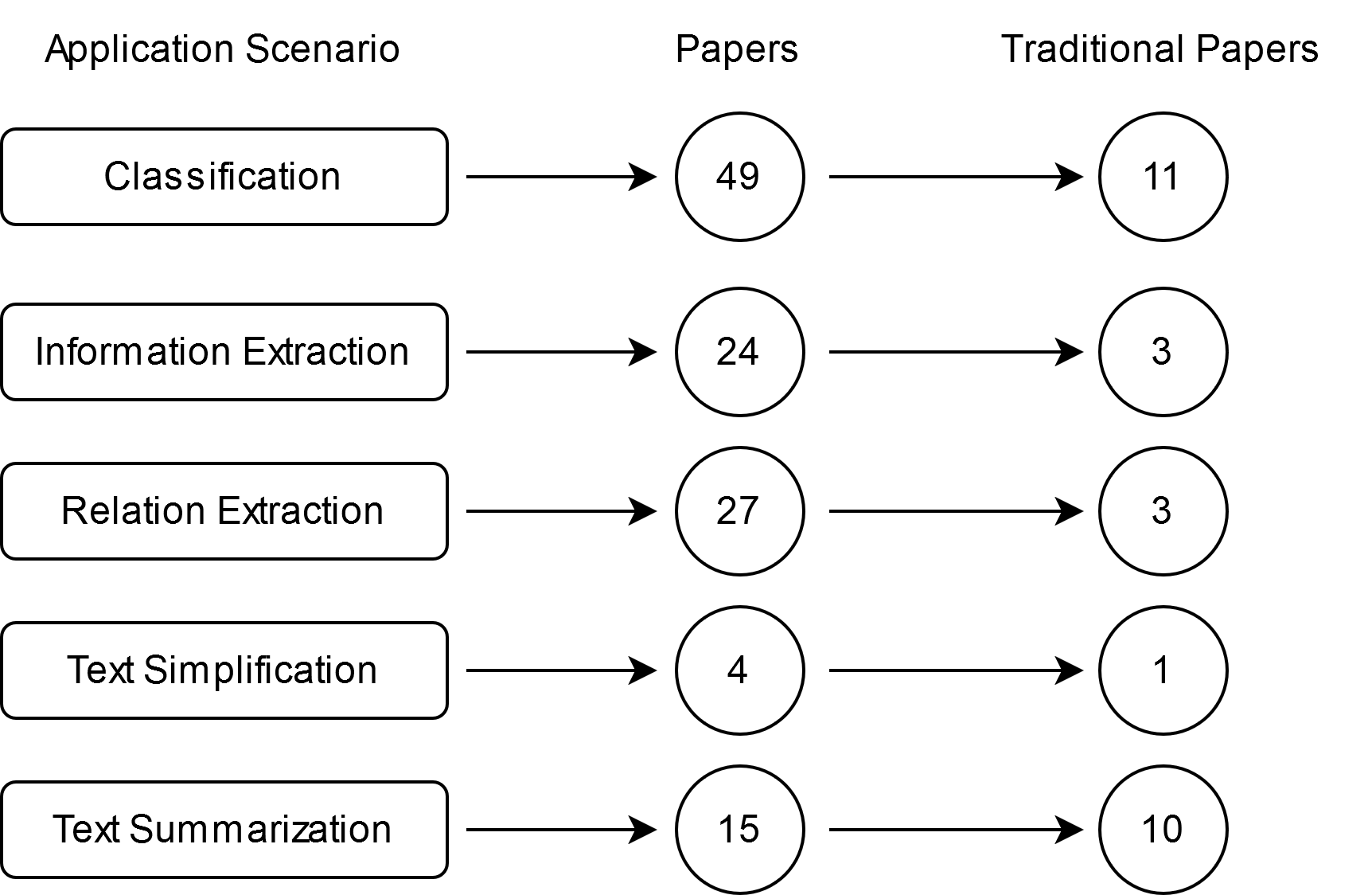}
\caption{Results of the SLR, ``Traditional Papers'' stands for the number of papers using traditional approaches in some way.}
\label{fig:slr}
\end{figure}

Therefore, the quantitative results for the SLR are displayed in figure~\ref{fig:slr}. Text classification was the dominant application scenario in terms of the total number of retrieved papers, followed by relation extraction, information extraction and text summarization. However, using the definition of traditional techniques applied in this paper, text summarization had the highest ratio of retrieved papers using\footnote{``using'' here ranges from applying as a benchmark over employing as a part of the pipeline to presenting as the main technique. This will be explained in more depth in the following section.} traditional approaches, being the only scenario with more than 50\% of results including such techniques. The high percentage of summarization papers with traditional papers is probably due to the more expansive definition of traditional approaches for this application scenario, including extractive techniques in general.

Thus,~\textbf{RQ1} can be answered positively, meaning for all application scenarios publications incorporating traditional models and techniques were retrieved within the confines of this SLR. The application scenario text simplification is the one outlier, necessitating the widening of search filters to include abstracts as well.

In terms of absolute numbers we have compared our results, retrieved from the ACM DL based on searches on the paper titles alone, with queries from the ACM DL on the abstracts of papers, as well as queries on Google Scholar on paper titles. Also, we filtered our results not only for papers published in 2023 but from 2015 up until the end of last year, 2024, as of January 27th 2025~\footnote{See the complete numbers on \url{https://doi.org/10.5281/zenodo.13683800}.}. The trends of classification being the largest application scenario are confirmed in both additional queries, as for simplification in being the smallest. Information extraction exceeds the numbers of both relation extraction and summarization scenarios, when looking at total numbers over the whole range of years.  Furthermore, Google Scholar retrieves roughly ten times as many papers as the ACM DL, due to the more strict peer review process performed by the ACM for a majority of its papers and the acceptance of pre-prints in Google Scholar.

As we laid out earlier, we have focused on 2023 due to the still ongoing publication processes for papers accepted for publication in 2024. However, in order to gauge the progress of NLP-research for 2024, we have updated our search in the ACM DL with the same search queries and did a smaller, more focused review, as a means to spotcheck, if the developments we have noticed for 2023 are still consistent in 2024. In terms of numbers, the amount of retrieved papers stays roughly the same year-over-year, with relation extraction differing the most, with 39 published papers in 2024. Due to size constraints, more complete numbers will be made available after the anonymity period ends.

Very briefly, papers from 2024 include traditional models, e.g., in \citep{pang2024research} by using SVMs as part of a pipeline or in \citep{sikosana2024comparative} with SVMs, Naive Bayes and Random Forest (RF) classifiers as comparison for more modern techniques, both papers for classification purposes. Extractive summarization approaches are still relevant as well, e.g., in \citep{zhou2023extractive} and \citep{gupta2024automatic}, while relation extraction, as comparison variants in \citep{shi2024agentre} and information extraction as part of the pipeline in \citep{liberatore2024quantitative} and as comparison in \citep{parfenova2024automating} both exhibit traditional techniques. Text simplification was adapted via rule-based techniques in case of syntactic simplification, in which the authors set themselves apart from current deep learning approaches \citep{keerthan2024syntactic}.

An overview of the qualitative results for the SLR can be found in table~\ref{tab:slr-description}.

\begin{table*}[!htpb]
\renewcommand{\arraystretch}{1.1} 
\centering
\begin{tabular}{@{}p{.04\textwidth}p{.93\textwidth}@{}}
\toprule
\multicolumn{2}{@{}l@{}}{\textbf{Classification}} \\
\hline
{[\citenum{chen_automatic_2024}]} & Linear n-gram classifier as one of three main models\\
{[\citenum{cunha_comparative_2023}]} & SVM as comparison technique, which achieves the best results on 2 datasets\\
{[\citenum{hong_protorynet_2023}]} & Decision trees, BoW and logistic regression models as comparison models\\
{[\citenum{lin_decision_2023}]} & Decision trees as main method as part of an assignment for a computer science course\\
{[\citenum{lu_simple_2022}]} & Combination of SVM and NB as baseline\\
{[\citenum{mehmood_enml_2023}]} & SVMs and NB as comparison benchmarks\\
{[\citenum{saeed_topic_2023}]} & Different traditional methods as main comparison methods\\
{[\citenum{shi_chinese_2023}]} & TF-IDF as part of pipeline\\
{[\citenum{wang_wot-class_2023}]} & SVM as pre-processing for BERT model\\
{[\citenum{yuan_joint_2022}]} & TF-IDF and point mutual information used for graph network construction in pipeline\\
{[\citenum{zhang_uncertainty_2023}]} & Gradient boosted decision trees as part of a tutorial\\
\multicolumn{2}{@{}l@{}}{\textbf{Information Extraction}} \\
\hline
{[\citenum{liu_information_2023}]} & SVM and keyword-methods as comparison methods\\
{[\citenum{sultana_extracting_2022}]} & 5 traditional models as core method of the paper (besides BERT for vector generation); application on news-paper articles, extracting 'sources' and classifying their type\\
{[\citenum{xiang_rule-based_2023}]} & Rule-based method as main model for extracting stock information from company news\\
\multicolumn{2}{@{}l@{}}{\textbf{Relation Extraction}} \\
\hline
{[\citenum{hu_selflre_2023}]} & Semi-supervised self-training as comparison baseline\\
{[\citenum{mazumder_graph_2023}]} & 6 traditional models as comparison baselines: SVM, logistic regression, RF, NB, link-based approach, LDA and Siamese networks; also uses IDF and point-wise mutual information as part of the pipeline\\
{[\citenum{osman_exploiting_2023}]} & SVM \& K-nearest neighbor as main classifiers for new hybrid approach\\
\multicolumn{2}{@{}l@{}}{\textbf{Text Simplification}} \\
\hline
{[\citenum{north_lexical_2023}]} & Classifiers for identifying lexical complexity: SVM, logistic regression, decision trees, RF
\\
\multicolumn{2}{@{}l@{}}{\textbf{Text Summarization}} \\
\hline
{[\citenum{atri_fusing_2023}]} & Extractive and traditional models as baseline and comparison to the main model\\
{[\citenum{bao_general_2023}]} & Combines abstractive \& extractive for rewriting; lead-3 as comparison\\
{[\citenum{dsilva_impact_2023}]} & Extends PageRank with different similarity measures (Cosine, Levenshtein Distance, FastText Distances); Lead-300 as comparison baseline\\
{[\citenum{du_extractive_2023}]} & Lead-3 \& TextRank as comparison baselines; use extractive approaches as a basis and add sequential decision-making and reinforcement learning\\
{[\citenum{lam_vietnamese_2022}]} & Segmentation of text into elementary discourse unit for abstractive \& extractive summaries\\
{[\citenum{licari_legal_2023}]} & LexRank as comparison baseline; use BERT models in the selection phase of sentences to rank them, and then use an extractive model in order to select sentences, thus hybrid model\\
{[\citenum{luo_gap_2022}]} & TextRank as main model, combined with gap sentences generation, thus hybrid model\\
{[\citenum{omar_method_2023}]} & Combines extractive \& statistical models; use semantic summary extractor; use LexRank \& TextRank as comparison baseline, and as part of the main model, thus hybrid model\\
{[\citenum{papagiannopoulou_social_2022}]} & SLR on social media summarization, 4 papers listed with extractive techniques\\
{[\citenum{park_acl_2023}]} & Pegasus model, uses abstractive \& extractive techniques on new dataset (ACL TA-DA)\\
\bottomrule
\end{tabular}
\caption{Results of the SLR. Only papers with traditional techniques are listed here, with a short description of the traditional technique and how each paper applies such techniques.}
\label{tab:slr-description}
\end{table*}

\section{Current Work on Traditional Approaches}
\label{sec:curr}

Three different usages of traditional approaches were observed in this SLR, i.e., traditional models as:
\begin{itemize}
    \item part of a pipeline for a larger, more complex model or approach, e.g.,~\citep{shi_chinese_2023,wang_wot-class_2023},
    \item comparison and/or baseline to the main model(s) of the paper, e.g.,~\citep{atri_fusing_2023,cunha_comparative_2023,hu_selflre_2023,liu_information_2023,mehmood_enml_2023}, or
    \item core method or one of the core methods of the paper, e.g.,~\citep{licari_legal_2023,osman_exploiting_2023,sultana_extracting_2022}.
\end{itemize}

For text classification, most observations regarding traditional techniques can be categorized as comparing approaches to the main topic or model of the paper. Techniques such as SVM, NB and linear $n$-gram classifiers were thus intended as a reference for other more modern techniques like BERT~\citep{lu_simple_2022} or BART~\citep{cunha_comparative_2023}. The latter paper~\citep{cunha_comparative_2023}, comprises experiments with various datasets and many models, including one traditional model based on SVM, which surpassed or matched the results of the other more modern models in three benchmarks, while displaying worse performance in many other benchmarks. 

Two papers retrieved for this application scenario presented the use of traditional models as part of an education task, first as an assignment of a computer science course~\citep{lin_decision_2023}, in which decision trees were used in order to present students an opportunity for a deeper understanding between the applied models and the data and even society itself. While another paper~\citep{zhang_uncertainty_2023} used traditional models in a tutorial regarding uncertainty in text classification, contrasting two main models, decision trees and BERT, as representatives of both traditional and modern techniques.

In one of the classification papers traditional models were used as core method(s), e.g., for evaluation of noise generation with a linear $n$-gram classifier as one of three main methods~\citep{chen_automatic_2024}, while traditional techniques such as TF-IDF were part of the pre-processing pipeline for two other approaches~\citep{shi_chinese_2023,yuan_joint_2022}.

The three retrieved papers for the information extraction scenario were either using traditional approaches as comparison method, with SVM and keyword extraction methods as reference for the main model~\citep{liu_information_2023}, or as core methods for two differing use-cases. Firstly, a selection of traditional models including SVM, RF and decision trees are used to extract source information from news texts~\citep{sultana_extracting_2022}. Secondly, company announcements are analyzed using a rule-based extraction method and a comparison of the rule-based method with machine learning and deep learning based approaches is done~\citep{xiang_rule-based_2023}.

For relation extraction, three articles with aspects of traditional approaches were found. In~\citep{hu_selflre_2023} the core method is compared with different methods including BERT, reinforcement learning techniques as well as semi-supervised self-training, where the last one can be categorized as a traditional approach. In a second paper, the main model, implemented using a GNN, is contrasted with various traditional models such as SVM and NB, however noting that SVM was the best baseline among the traditional models and even beating their proposed model in one benchmark, while trailing behind in all other experiments~\citep{mazumder_graph_2023}. Thirdly, SVM and $k$-nearest-neighbor methods are used as main classifiers in order to extract complex Arabic relations as part of a new machine learning based hybrid approach, thus using a traditional model as core method~\citep{osman_exploiting_2023}.

Text simplification in terms of traditional approaches is even more constrained since only one paper could be retrieved. There, a survey on lexical complexity prediction is done, including techniques such as SVMs, decision trees, RF classifiers and modern approaches~\citep{north_lexical_2023}.

Lastly, the papers retrieved for text summarization comprise ten papers with extractive techniques as traditional approaches. Since many extractive models with the characteristics mentioned above in section~\ref{sec:term} are still relevant and in use, as shown in these ten papers, the loose definition is still fitting. Extractive techniques, sequential decision-making and reinforcement learning are adapted, using traditional extractive systems as comparison in the  evaluation~\citep{du_extractive_2023}. Prominent extractive approaches such as LexRank~\citep{erkan2004lexrank} and TextRank~\citep{mihalcea2004textrank} are mentioned multiple times, as comparison for other models, in which limitations in both extractive and abstractive summarization approaches are noted, or as part of the main model, thus a hybrid model~\citep{atri_fusing_2023,omar_method_2023}. 

Other hybrid approaches use BERT models as part of the ranking phase before selecting the sentences through a purely extractive summarization approach~\citep{licari_legal_2023}. The approach, in this case, is chosen due to the strict nature of the domain, which comprises legal texts. For less restrictive use-cases, abstractive and extractive approaches are combined as a pre-processing step in order to compose a summary, while also comparing the generated text with a summary produced by lead-3, itself being an even older approach than the previously mentioned extractive methods~\citep{bao_general_2023}. More hybrid approaches use Text\-Rank and Gap Sentences Generation~\citep{luo_gap_2022}, extending the prior hybrid model PEGASUS~\citep{zhang2020pegasus}. PEGASUS was also used as the main model in a combined effort of abstractive and extractive techniques as part of the pipeline~\citep{park_acl_2023}. Many summarization approaches listed here apply extractive techniques as part of the main model or even as the main method~\citep{du_extractive_2023,licari_legal_2023,luo_gap_2022,omar_method_2023}.

It has to be noted that some approaches reviewed here are narrow or in a confined application area, e.g., in low-resource languages~\citep{dsilva_impact_2023} or domains~\citep{hu_selflre_2023}. 

As for~\textbf{RQ2}, we can again confirm the existence of traditional models in the retrieved papers and also their usage in a myriad of ways. 
While most papers use traditional approaches in the pipelines or as comparison, a few selected papers were determined that use traditional models as one of or even the one core approach of the paper. A summary regarding the usage type of each paper listed in this section is displayed in table~\ref{tab:slr-categories}. Two table entries are listed without checkmarks in the columns on the right-hand side, which both are survey papers without new approaches or without a quantitative comparison between models ~\citep{north_lexical_2023,papagiannopoulou_social_2022}.

\section{Application Scenarios for Traditional Approaches}
\label{sec:appsec}

After reviewing the results of the SLR some scenarios will be discussed, which will show the continued relevance of traditional approaches in NLP.

As mentioned above in section~\ref{sec:curr}, rule-based methods were used as the main model for information extraction, in which information on stock developments are identified based on company announcements~\citep{xiang_rule-based_2023}. Similarly, in relation extraction, entity relations in Arabic texts are retrieved~\citep{osman_exploiting_2023}. Both approaches can be categorized as a rather narrow task, where hand-crafted rules can be manually written. 


In case of text summarization, extractive approaches are not only more efficient due to their simple nature, but also cannot include hallucinations or even sentence constructions that do not appear in the original texts. In applications, where lexical, grammatical or syntactic errors and hallucinations must not appear, extractive summarization approaches are thus an obvious choice. The legal domain is one sector in which such requirements are apparent, as mentioned in section~\ref{sec:curr}. One example is a summarization approach for Italian legal texts~\citep{licari_legal_2023}. In this hybrid system the ranking is done by a BERT model, which feeds into the selection model, implemented as an extractive system using the LexRank algorithm.


When looking back at~\textbf{RQ3}, a direct analysis or discussion on the advantages of traditional models compared to other more modern state-of-the-art models was not included in the retrieved papers, apart from quantitative evaluations comparing the selected metrics. However, a few advantages have become obvious through the selection of the traditional models in three papers. Firstly, in the summarization scenario, an extractive model was used in order to prevent hallucinations from appearing~\citep{licari_legal_2023} or, secondly, in terms of explainability and reproducibility in the scenarios information and relation extraction to comprehend and track the sequences of operation~\citep{osman_exploiting_2023,xiang_rule-based_2023}.

\begin{table}[H]
\renewcommand{\arraystretch}{1.18} 
\centering
\begin{tabular}{@{}lcccc@{}}
\toprule
Paper & Pipeline & \multicolumn{2}{@{}c@{}}{Comparison} & Core \\
\cmidrule{3-4} &  & Compe. & Tr. Be. &  \\
\midrule
\multicolumn{2}{@{}l@{}}{\textbf{Classification}} \\
\hline
{[\citenum{chen_automatic_2024}]} & & \checkmark  && \checkmark  \\
{[\citenum{cunha_comparative_2023}]} & \checkmark & \checkmark & \checkmark &\\
{[\citenum{hong_protorynet_2023}]} & & \checkmark && \\
{[\citenum{lin_decision_2023}]} & &&& \checkmark \\
{[\citenum{lu_simple_2022}]} & && \checkmark &\\
{[\citenum{mehmood_enml_2023}]} &&& \checkmark &\\
{[\citenum{saeed_topic_2023}]} & & \checkmark & &\\
{[\citenum{shi_chinese_2023}]} & \checkmark &&&\\
{[\citenum{wang_wot-class_2023}]} & \checkmark &&&\\
{[\citenum{yuan_joint_2022}]} & \checkmark &&&\\
{[\citenum{zhang_uncertainty_2023}]} & &&& \checkmark \\
\hline
\multicolumn{2}{@{}l@{}}{\textbf{Information Extraction}} \\
\hline
{[\citenum{liu_information_2023}]} &&& \checkmark &\\
{[\citenum{sultana_extracting_2022}]} & &&& \checkmark \\
{[\citenum{xiang_rule-based_2023}]} & & \checkmark && \checkmark\\
\hline
\multicolumn{2}{@{}l@{}}{\textbf{Relation Extraction}} \\
\hline
{[\citenum{hu_selflre_2023}]} & && \checkmark &\\
{[\citenum{mazumder_graph_2023}]} & \checkmark & \checkmark & \checkmark &\\
{[\citenum{osman_exploiting_2023}]} && \checkmark && \checkmark \\
\hline
\multicolumn{2}{@{}l@{}}{\textbf{Text Simplification}} \\
\hline
{[\citenum{north_lexical_2023}]} & &&&\\
\hline
\multicolumn{2}{@{}l@{}}{\textbf{Text Summarization}} \\
\hline
{[\citenum{atri_fusing_2023}]} & && \checkmark &\\
{[\citenum{bao_general_2023}]} & \checkmark && \checkmark &\\
{[\citenum{dsilva_impact_2023}]} & \checkmark & \checkmark && \checkmark\\
{[\citenum{du_extractive_2023}]} & && \checkmark &\\
{[\citenum{lam_vietnamese_2022}]} & \checkmark & \checkmark &&\\
{[\citenum{licari_legal_2023}]} & \checkmark & \checkmark && \checkmark \\
{[\citenum{luo_gap_2022}]} & & \checkmark && \checkmark \\
{[\citenum{omar_method_2023}]} & &&& \checkmark \\
{[\citenum{papagiannopoulou_social_2022}]} &&&&\\
{[\citenum{park_acl_2023}]} & \checkmark &&&\\
\bottomrule
\end{tabular}
\caption{Results of the SLR with categorization if the traditional technique is part of the processing pipeline (column 2) or part of a comparison with the main models of the paper (columns 3 \& 4). Column 3 indicates papers that include experiments, in which traditional models are competitive with modern approaches (``Compe.''), whereas column 4 displays papers, in which traditional models trail behind modern approaches (``Tr. Be.''). Papers marked in column 5 use traditional approaches as core model(s).}
\label{tab:slr-categories}
\end{table}

\section{Conclusion and Future Work}
\label{sec:concl}

Even though the impact of LLMs cannot be overstated for the research field of NLP, the role of traditional models, approaches and techniques is still apparent, which was shown in this SLR. After shortly presenting the state of the art for the application areas determined for this SLR, the terminology and key factors characterizing traditional models, which include reproducibility, efficiency and more, were presented. Through a literature review of publications listed in the ACM Digital Library from 2023 we have determined that for all five applications scenarios that were analyzed in this SLR, references, applications or even continued research through and with traditional models can be found.

Text classification and text summarization scenarios included, in terms of quantitative results, the majority of traditional approaches, mostly comprising support vector machines, Naive-Bayes classifiers or extractive approaches. But other techniques such as rule-based approaches or decision trees, although not in a similar scope as for classification and summarization, were retrieved for information and relation extraction as well as for text simplification. Furthermore, hybrid systems using traditional aspects in parts of the system were found in multiple publications, and other papers still use traditional approaches as point of comparison for the main model.

When reviewing our initial research questions,~\textbf{RQ1} can be answered affirmatively, since 28 of the 119 papers retrieved in this SLR incorporate traditional models. Regarding \textbf{RQ2}, different purposes for traditional models and techniques were identified for all five application scenarios, such as being part of the processing pipelines, as comparison or baseline for the main models, or even as the main model or models of some papers (typically for specific domains). As for~\textbf{RQ3}, no direct discussion was found in the papers identified in this SLR. However, indirect comments regarding reproducibility, explainability and removing uncertainty when dealing with LLMs, especially preventing hallucinations from appearing, were noted.


In our future work, we will further look at traditional models and work on criteria and conditions that suggest the use of traditional approaches in NLP. We will also identify and characterize example scenarios in which their use should be considered. 

\section{Limitations}
\label{sec:limit}
We have focused our survey on a selection of application scenarios, namely classification, information extraction, relation extraction, text simplification and text summarization. The five tasks offer a broad spectrum of different NLP use-cases, however an over-arching view across the whole NLP field is not possible, even with this range of tasks. Still, within these five tasks, we include some tasks that can be categorized as extraction of targeted information or classification tasks, while others produce new text based on the input, in the case of simplification and summarization.

Also, we have focused on 2023 as the main year for our literature review. We have excluded older publications, since we wanted to gain an overview on the NLP research field and if traditional models, as defined in section \ref{sec:term}, can still be found in the literature. 
In addition to the review of literature published in 2023, we have also included a brief review of further developments in 2024 in section \ref{sec:slr}. However, an SLR like this one can of course only ever be a snapshot in time. Nevertheless, we have gained the impression that at least some of the reasons for the continued consideration of traditional methods will remain valid in the longer term.

Another restriction of our work is the focus on the ACM DL. We acknowledge this restriction, however our aim from the start was not an all-encompassing view on current NLP-research, but a focused investigation, on a selected database with high-quality publications for a recent time frame and a plausible sample of current NLP research.

Lastly, our definition for ``traditional models'' could be scrutinized. We employed a pragmatic distinction depending on the task at hand, because a clear-cut definition is difficult, if not impossible. Therefore, we included the explicit section on terminology in section \ref{sec:term}. Also, due to page constraints, we did not offer a discussion on the middle-ground between traditional models and modern deep learning approaches, which could be roughly seen as early embeddings spaces such as word2vec. 


\setcitestyle{numbers}
\bibliography{acl_latex}

\end{document}